\documentclass{article} 
\usepackage{iclr2015,times}
\usepackage{hyperref}
\usepackage{graphicx,graphics,wrapfig,epstopdf,subfig}
\usepackage{url}
\usepackage{graphicx}
\usepackage{amsmath}
\usepackage{amssymb}
\newcommand{\Dmat}{{\bf D}}

\newcommand{\Xmat}{{\bf X}}

\iclrfinalcopy 

\title{A Generative Model for Deep Convolutional Learning}

\author{
Yunchen Pu, Xin Yuan and Lawrence Carin \\
Department of Electrical and Computer Engineering, Duke University,
Durham, NC, 27708, USA \\
\texttt{\{yunchen.pu,xin.yuan,lcarin\}@duke.edu} 
}
\begin{document}

\maketitle

\begin{abstract}
A generative model is developed for deep (multi-layered) convolutional dictionary learning.
A novel {probabilistic pooling} operation is integrated into the deep model, yielding efficient bottom-up (pretraining) and top-down (refinement) probabilistic learning. 
Experimental results demonstrate powerful capabilities of the model to learn multi-layer features from images, and excellent classification results are obtained on the MNIST and Caltech 101 datasets. 
\end{abstract}

\section{Introduction}
We develop a deep generative statistical model, which starts at the highest-level features, and maps these through a sequence of layers, until ultimately mapping to the data plane (e.g., an image). The feature at a given layer is mapped via a multinomial distribution to one feature in a block of features at the layer below (and all other features in the block at the next layer are set to zero). 
This is analogous to the method in~\cite{Lee09ICML}, in the sense of imposing that there is {\em at most} one non-zero activation within a pooling block. 
We use bottom-up pretraining, in which initially we sequentially learn parameters of each layer one at a time, from bottom to top, based on the features at the layer below. However, in the refinement phase, all model parameters are learned jointly, top-down. Each consecutive layer in the model is locally conjugate in a statistical sense, so learning model parameters may be readily performed using sampling or variational methods.

\vspace{-3mm}
\section{Modeling Framework}
Assume $N$ gray-scale images $\{{\Xmat^{(n)}}\}_{n=1,N}$, with $\Xmat^{(n)}\in\mathbb{R}^{N_x \times N_y}$; the images are analyzed jointly to learn the convolutional dictionary $\{{\Dmat^{(k)}}\}_{k=1,K}$. Specifically
consider the model 
\vspace{-2mm}
\begin{equation}\label{Eq:betabern}
\Xmat^{(n)} = \sum_{k=1}^K  \Dmat^{(k)} \ast ({\bf Z}^{(n,k)}\odot {\bf W}^{(n,k)})  + {\bf E}^{(n)},
\vspace{-2mm}
\end{equation}
where $\ast$ is the convolution operator, $\odot$ denotes the Hadamard (element-wise) product, the elements of ${\bf Z}^{(n,k)}$ are in $\{0,1\}$, the elements of ${\bf W}^{(n,k)}$ are real, and ${\bf E}^{(n)}$ represents the residual. ${\bf Z}^{(n,k)}$ indicates which shifted version of ${\bf D}^{(k)}$ is used to represent ${\bf X}^{(n)}$. 

Assume an $L$-layer model, with layer $L$ the top layer, and layer 1 at the bottom, closest to the data. In the pretraining stage, the output of layer $l$ is the input to layer $l+1$, after pooling. 
Layer $l\in\{1,\dots,L\}$ has $K_l$ dictionary elements, and we have:
\begin{eqnarray}
{\bf X}^{(n, l+1)} &=& \textstyle \sum_{k_{l+1}=1}^{K_{l+1}} {\bf D}^{(k_{l+1}, l+1)} * \left({\bf Z}^{(n,k_{l+1},l+1)}  \odot {\bf W}^{(n,k_{l+1}, l+1)}\right) + {\bf E}^{(n, l+1)} \label{Eq:x_lp1}\\
{\bf X}^{(n, l)} &=& \textstyle \sum_{k_{l}=1}^{K_{l}} {\bf D}^{(k_{l}, l)} * \underbrace{\left({\bf Z}^{(n,k_{l},l)} \odot {\bf W}^{(n,k_{l}, l)}\right)}_{= {\bf S}^{(n,k_{l},l)}} + {\bf E}^{(n, l)} \label{Eq:x_l}
\end{eqnarray}
The expression ${\bf X}^{(n,l+1)}$ may be viewed as a 3D entity, with its $k_l$-th plane defined by a ``pooled'' version of ${\bf S}^{(n,k_l,l)}$.  

\begin{figure}[tbp!]
	\centering
	\vspace{-3mm}
	\includegraphics[scale=0.3]{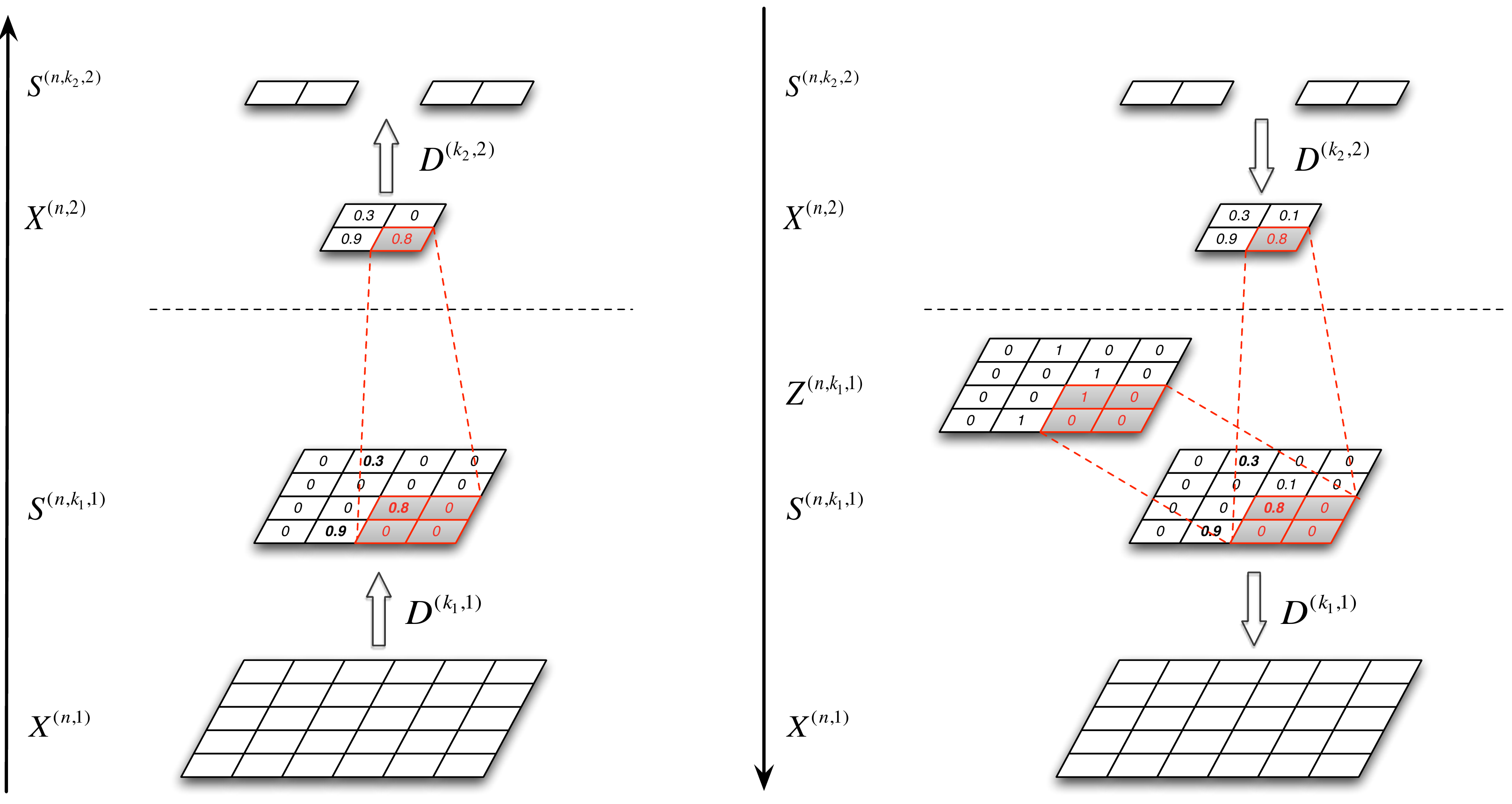}
	\vspace{-3mm}
	\caption{\small{Schematic of the proposed generative process. Left: bottom-up pretraining, right: top-down refinement. (Zoom-in for best visulization and a larger version can be found in the Supplementary Material.)}}
	\vspace{-6mm}
	\label{fig:max_pool}
\end{figure}

The 2D activation map ${\bf S}^{(n,k_l,l)}$ is partitioned into $n_x\times n_y$ dimensional contiguous blocks (pooling blocks with respect to layer $l+1$ of the model); see the left part of Figure \ref{fig:max_pool}. Associated with each block of pixels in ${\bf S}^{(n,k_l,l)}$ is one pixel at layer $k_l$ of ${\bf X}^{(n,l+1)}$; the relative locations of the pixels in ${\bf X}^{(n,l+1)}$ are the same as the relative locations of the blocks in ${\bf S}^{(n,k_l,l)}$. Within each block of ${\bf S}^{(n,k_l,l)}$, either all $n_xn_y$ pixels are zero, or only one pixel is non-zero, with the position of that pixel selected stochastically via a multinomial distribution. Each pixel at layer $k_l$ of ${\bf X}^{(n,l+1)}$ equals the largest-amplitude element in the associated block of ${\bf S}^{(n,k_l,l)}$ ($i.e.$, max pooling).

%

The learning performed with the top-down generative model (right part of Fig.~\ref{fig:max_pool}) constitutes a {\em refinement} of the parameters learned during pretraining, and the excellent initialization constituted by the parameters learned during pretraining is key to the subsequent model performance.

In the refinement phase, we now proceed top down, from (\ref{Eq:x_lp1}) to (\ref{Eq:x_l}). The generative process constitutes ${\bf D}^{(k_{l+1}, l+1)}$ and ${\bf Z}^{(n,k_{l+1},l+1)}  \odot {\bf W}^{(n,k_{l+1}, l+1)}$, and after convolution ${\bf X}^{(n,l+1)}$ is manifested; the ${\bf E}^{(n,l)}$ is now absent at all layers, except layer $l=1$, at which the fit to the data is performed. Each element of ${\bf X}^{(n,l+1)}$ has an associated pooling {\em block} in ${\bf S}^{(n,k_l,l)}$. 

\vspace{-4mm}
\section{Experimental Results}
\label{Sec:Exp}
\vspace{-4mm}
We here apply our model to the MNIST and Caltech 101 datasets.
\vspace{-3mm}
\paragraph{MNIST Dataset}
\begin{wraptable}{r}{0.50\textwidth}
	\vspace{-6mm}
	\caption{\small{Classification Error of MNIST data}}
	\vspace{-4mm}
	\centering
		\small
	\begin{tabular}{c|c}
		Methods & Test error \\
		\hline 
		$\begin{array}{l}
		\text{6-layer Conv. Net + 2-layer Classifier } \\
		\text{+ elastic distortions~\cite{Ciresan11IJCAI}}
		\end{array}$
		& 0.35\% \\
		\hline
			MCDNN~\cite{ciresan2012multi} & 0.23\%\\
		\hline
		SPCNN~\cite{Zeiler13ICLR} & 0.47\%\\
		\hline
		$\begin{array}{l}
		\text{HBP~\cite{Chen13deepCFA},}\\
		\text{2-layer cFA + 2-layer features} \end{array}$& 0.89\% \\
		\hline
		Ours, 2-layer model + 1-layer features &0.42\%  \\	
	\end{tabular}
	\label{Table:Error_MNIST}
	\vspace{-6mm}
\end{wraptable}
%

Table~\ref{Table:Error_MNIST} summaries the classification results of our model compared with some related results, on the MNIST data.
The second (top) layer features corresponding to the refined dictionary are sent to a nonlinear
support vector machine (SVM)~\citep{CC01a} with Gaussian kernel, in a one-vs-all multi-class classifier, with classifier parameters tuned via 5-fold cross-validation (no tuning on the deep feature learning).

\vspace{-3mm}
\paragraph{Caltech 101 Dataset}

\begin{wraptable}{r}{0.53\textwidth}
\vspace{-3mm}
	\caption{ \small{Classification Accuracy Rate of Caltech-101.}}
	\vspace{-0.3cm}
	\centering
	\small
	\begin{tabular}{c|c|c}
		\# Training Images per Category & 15 & 30 \\
		\hline
		DN~\cite{Zeiler10CVPR}  & 58.6 \% & 66.9\% \\
		\hline
		CBDN~\cite{Lee09ICML} & 57.7 \% & 65.4\% \\
		\hline
		HBP ~\cite{Chen13deepCFA}  & 58\% & 65.7\% \\
		\hline
		ScSPM ~\cite{yang09CVPR} & 67 \% & 73.2\%  \\
		\hline
		P-FV ~\cite{seidenari2014local} & 71.47\% & 80.13\% \\
		\hline
		R-KSVD ~\cite{li2013reference}  & 79 \% & 83\% \\
		\hline
		Convnet~\cite{Zeiler14ECCV}  & 83.8 \% & 86.5\% \\	
		\hline
		Ours, 2-layer model + 1-layer features  & 70.02\% & 80.31\%  \\
		\hline
		Ours, 3-layer model + 1-layer features & 75.24\% & 82.78\% 
	\end{tabular}
	\label{Table:accuracy_caltech101}
	\vspace{-0.4cm}
\end{wraptable}
We next consider the Caltech 101 dataset.For Caltech 101 classification, we follow the setup in~\cite{yang09CVPR}, selecting 15 and 30 images per category for training, and testing on the rest.
The features of testing images are inferred based on the top-layer dictionaries and sent to a multi-class SVM; we again use a Gaussian kernel non-linear SVM with parameters tuned via cross-validation.
Ours and related results are summarized in Table~\ref{Table:accuracy_caltech101}.
\vspace{-3mm}
\section{Conclusions}
\vspace{-4mm}
A deep generative convolutional dictionary-learning model has been developed within a Bayesian setting. The proposed framework enjoys efficient bottom-up and top-down probabilistic inference. A probabilistic pooling module has been integrated into the model, a key component to developing a principled top-down generative model, with efficient learning and inference.
Extensive experimental results demonstrate the efficacy of the model to learn multi-layered features from images. 

\bibliography{iclr2015}

\end{document}